\documentclass[conference]{IEEEtran}
\usepackage{times}
\usepackage{soul}
\usepackage{url}
\usepackage{amsfonts}
\usepackage[hidelinks]{hyperref}
\usepackage[utf8]{inputenc}
\usepackage[small]{caption}
\usepackage{graphicx}
\usepackage{amsmath}
\usepackage{amsthm}
\usepackage{booktabs}
\usepackage{algorithm}
\usepackage{algorithmic}
\usepackage{amsmath}
\usepackage{amssymb}
\usepackage{subcaption}
\usepackage{tabularx}
\usepackage{multirow}
\usepackage[bottom]{footmisc}

\newcommand{\pmemwithoutspace}{Optane-PM}
\newcommand{\pmem}{\pmemwithoutspace }
\newcommand{\devdax}{Dev-DAX }
\newcommand{\fsdax}{FS-DAX }

\title{Non-Volatile Memory Accelerated Posterior Estimation}

\author{
\IEEEauthorblockN{Andrew Wood}
\IEEEauthorblockA{Boston University \\
aewood@bu.edu}

\and

\IEEEauthorblockN{Moshik Hershcovitch}
\IEEEauthorblockA{IBM Research \\
moshikh@il.ibm.com}

\and

\IEEEauthorblockN{Daniel Waddington}
\IEEEauthorblockA{IBM Research\\
daniel.waddington@ibm.com}

\and 

\IEEEauthorblockN{Sarel Cohen}
\IEEEauthorblockA{Hasso Plattner Institute \\ 
sarel.cohen@hpi.de}

\and

\IEEEauthorblockN{Peter Chin}
\IEEEauthorblockA{Boston University \\
spchin@bu.edu}
}

\begin{document}
\maketitle

\begin{abstract}
Bayesian inference allows machine learning models to express uncertainty. Current machine learning models use only a single learnable parameter combination when making predictions, and as a result are highly overconfident when their predictions are wrong. To use more learnable parameter combinations efficiently, these samples must be drawn from the posterior distribution. Unfortunately computing the posterior directly is infeasible, so often researchers approximate it with a well known distribution such as a Gaussian. In this paper, we show that through the use of high-capacity persistent storage, models whose posterior distribution was too big to approximate are now feasible, leading to improved predictions in downstream tasks.
\end{abstract}

\section{Introduction}
Machine learning models, particularly neural networks, are prone to overfitting. As a result, models are overconfident in their predictions, especially when those predictions are incorrect. This can have devastating consequences for agents acting on those predictions; such as autonomous cars or medical diagnosis.

Representing uncertainty has been studied for decades~\cite{blundell2015weight,chen2014stochastic,draper1995assessment,hansen2001robust}. It is well known that the lack of uncertainty is an artifact of learning a point mass estimate on incomplete data~\cite{draper1995assessment,hansen2001robust,kingma2015variational}. For instance, if all possible data points could be collected, then the optimal learning model would overfit to that data. However, because of finite data, models which learn a single parameter combination cannot make optimal predictions.

To make optimal predictions, researchers have modeled the learning setting from a Bayesian perspective. When learning, a model seeks to find the most probable parameter combination given the data $P(\theta | D)$. To then make optimal predictions, one must marginalize out the parameters. To do this, one must first compute the posterior distribution $P(y | \theta, x)$. Predictions can then be made by the following equation:
$$Pr(y | x) = \int Pr(y | \theta, x) Pr(\theta | D) d\theta$$
However, learning the posterior is not trivial. The posterior is given from Bayes' rule:
$$Pr(\theta | D) = \frac{Pr(D | \theta)Pr(\theta)}{Pr(D)}$$
where $Pr(D|\theta)$ is given via classical point estimate algorithms such as stochastic gradient descent. The trouble lies with computing $Pr(D)$. This term could of course be conditioned and then marginalized:
$$Pr(D) = \int Pr(D|\theta)Pr(\theta)d\theta$$
however this would require computing every possible parameter combination: an infeasible task.

Therefore, researchers have often turned to approximating the posterior using well known distributions. One fruitful avenue of research has been approximating with a multivariate gaussian~\cite{maddox2019simple}. However, posterior approximations require the parameters of the distribution be stored and updated as the model trains. For gaussians in particular, the computational demand is intense. To represent a $n$-dimensional multivariate gaussian, a covariance matrix of size $O(n^2)$ must be stored.

Until recently, there were only two ways of manipulating data on a von-Neumann architecture: volatile system memory (DRAM) and memory mapping (MMAP) data to disk. While DRAM is fast, it is not high capacity. On the other hand, MMAP storage is high capacity, but slow. To compute medium to large size posteriors, MMAP was the only method that dramatically increased the runtime of training.

In the last few years, a notable hardware breakthrough has been the emergence of Intel Optane Persistent Memory Modules (\pmemwithoutspace). \pmem in particular communicates via the memory bus, circumventing bottlenecks such as PCI-express lane availability, using the same interface to the CPU as DRAM. While there are other Persistent Memory technologies, \pmem is the most mature product on the market. \pmem is based on 3D-XPoint (3DXP) technology and operates at a cache-line granularity with a latency of around 300ns~\cite{spectra2020, izraelevitz2019basic}. While this latency is slower than current DRAM ({\raise.17ex\hbox{$\scriptstyle\mathtt{\sim}$}}100ns), it is 30x faster than the current state of the art NVMe SSDs. Additionally, a single DIMM of \pmem can reach 512GB, which is 8x larger than the available DRAM.  Thus, the maximum \pmem capacity of a commodity 2U server machine is 12TB - significantly more than DRAM.

In this paper, we show that by using \pmemwithoutspace, existing posterior approximation techniques can extend to models that could be previously handled due to memory and speed constraints. We demonstrate this using approximations that require six to 470 GB of storage trained on the MNIST dataset. We compare our results against approximating the posterior in DRAM and traditional memory mapping. 

\section{Experiments}
In our experiments, we operate on the well known MNIST dataset~\cite{lecun1998mnist}. MNIST is a popular benchmark for a variety of reasons: it is well curated, the complexity of the problem is low, and it is small. We chose this dataset because of the low problem complexity: we wish to test our implementation on a posterior which, while complex, is reasonable to expect models to learn.

When estimating the posterior, the size of the data does not affect the storage or the runtime of the approximation algorithm. The approximation algorithm is instead entirely defined by the number of learnable parameters the model contains. In our case of using a gaussian, the approximation scales linearly with the number of parameters in the model. For the gaussian to be full-rank, we will need to store $|\theta|$ separate parameter vectors sampled from the SGD trajectory. Following Maddox~\emph{et al}~\cite{maddox2019simple}, parameter vectors are used to compute the columns of a dense matrix $\hat{D}$. $\hat{D}$ is then used to estimate off-diagonal entries of the covariance during sampling. $\hat{D}$ induces a $O(T|\theta|)$ storage cost where $T$ is the rank of the approximation. The storage cost of this matrix explodes as it approaches full-rank: requiring $O(|\theta|^2)$ memory.

\begin{table}[!t]
    \centering
    \begin{tabular}{|c|c|c|c|}
        \hline
        Name & Management & Persistent & Speed/DRAM ratio \\
        \hline
        \fsdax & filesystem & yes & 3x \\
        \devdax & direct device & yes & 3x\\
        Memory Mode & as DRAM & no & 1x \\ 
        \hline
    \end{tabular}
    \caption{The three configurations of \pmemwithoutspace. Note that in Memory Mode, \pmem is only accessible through DRAM cache misses. Both \fsdax and \devdax offer cache-line granularity with direct load/store access. Memory mode requires pairing a DRAM module with an \pmem module on the same memory channel. The CPU's memory controller manages the cache transparently.}
    \label{tab:pmem_config}
    \vspace{-5mm}
\end{table}

\begin{table}[!b]
    \vspace{-4mm}
    \begin{tabular}{|c|c|c|c|c|}
        \hline
        & 1 epoch & 25 epochs & 50 epochs & 75 epochs\\
        \hline
        Size (GB) & 6.28 & 156.33 & 312.62 & 468.92 \\
        Approximation Rank      & 600  & 15k    & 30k    & 45k \\
        \hline
    \end{tabular}
    \caption{Posterior size for our four layer fully connected model. Note that the posterior size is controlled by the number of samples recorded (and hence the number of epochs during training). Note that even with the high storage cost, our approximation is still low-rank (full-rank = 2.8M).}
    \label{table:posterior_sizes}
\end{table}

We chose a model consisting of four fully connected layers as it is simple enough to learn MNIST while being large enough to compare \pmem to traditional storage methods. Our model contains 2.8M parameters, which requires 11MB of memory. During training, we treat the learnable parameters after every minibatch as a sample drawn from the posterior and use it to update the corresponding gaussian~\cite{maddox2019simple}. In our experiments, we store one posterior approximation entirely on DRAM, another on a memory-mapped filesystem, and the other on \pmemwithoutspace. We make use of a Python library called PyMM~\cite{WaHeDi21PyMM}\footnote{https://github.com/IBM/pymm}. PyMM is a specialized version of a software framework called Memory Centric Active Storage (MCAS)~\cite{waddington2021architecture} which removes the network layer and only uses local storage devices. MCAS is a key-value store built from the ground up that provides an interface between a client application and \pmemwithoutspace. Data that is stored via PyMM is persistent and can be operated on in-place; meaning code operates directly on the device without requiring a copy or transfer to DRAM. We evaluate \pmem using three modes as shown in table~\ref{tab:pmem_config}.

We measure the runtime of learning the posterior approximation three times per storage method, and measure it as a function of the number of training epochs. During each epoch, we update the posterior after every minibatch (600 minibatches per epoch with a minibatch size of 100). The memory size of our posterior can be seen in Table~\ref{table:posterior_sizes}.

Our experiments were conducted on Lenovo SR650 2U server equipped with two Intel Xeon Gold 6248 (2.5GHz) processors supporting 80 CPU hardware threads. The server is also equipped with 384GB (12x32GB) of DDR4 DRAM and 1.5TB of \pmem (12x128GB).
Our platform also includes an NVIDIA Tesla M60 GPU, which we used to perform traditional stochastic gradient descent calculations. Note that PyMM can transfer memory directly to all devices attached to the CPU, meaning that we transfer memory to and from the GPU and \pmem without first copying to DRAM.

\begin{figure}[!t]
\vspace{-8mm}
    \includegraphics[width=\columnwidth,trim=1 1 2 1,clip]{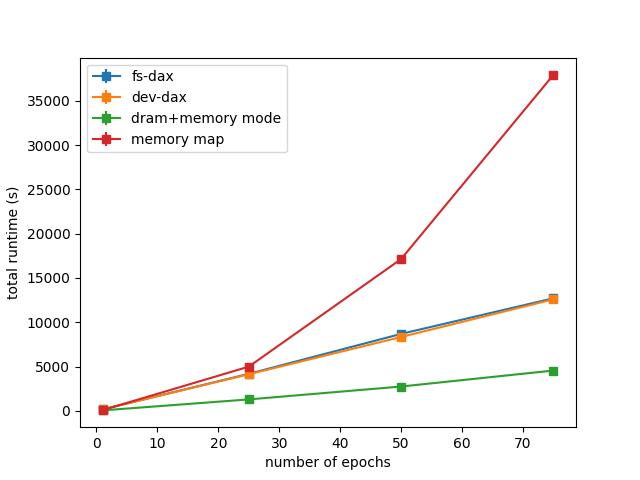}
\vspace{-1mm}
    \caption{\textbf{Total runtime of MNIST training}. The posterior is updated every minibatch. Note that memory mapping does not scale with the posterior size (controlled via the number of training epochs).}
\vspace{-5mm}
    \label{fig:total_runtimes}
\end{figure}

\begin{figure}[!b]
\vspace{-8mm}
    \includegraphics[width=\columnwidth,trim=1 1 2 1,clip]{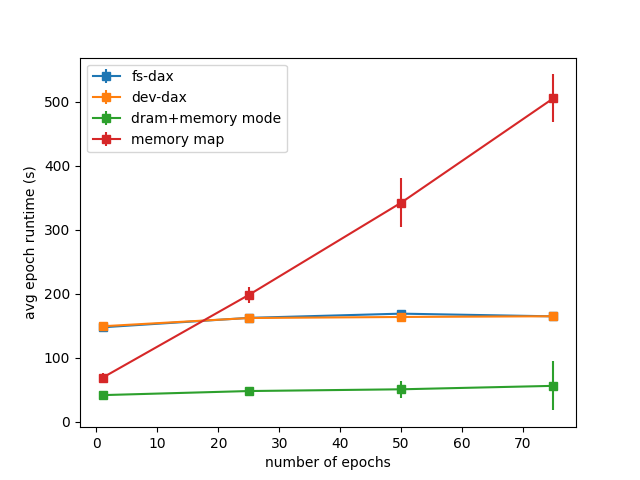}
\vspace{-1mm}
    \caption{\textbf{Avg epoch runtime of MNIST training} with standard deviation. Note the high variance in Memory mode where the posterior is beyond DRAM capacity. This is due to the random access nature of updating the posterior and processing cache misses/evictions.}
\vspace{-5mm}
    \label{fig:avg_epoch_runtimes}
\end{figure}

\section{Results and Discussion}
\begin{figure}[!t]
\vspace{-8mm}
    \hspace{-4.4mm}\includegraphics[width=1.2\columnwidth,trim=1 1 2 1,clip]{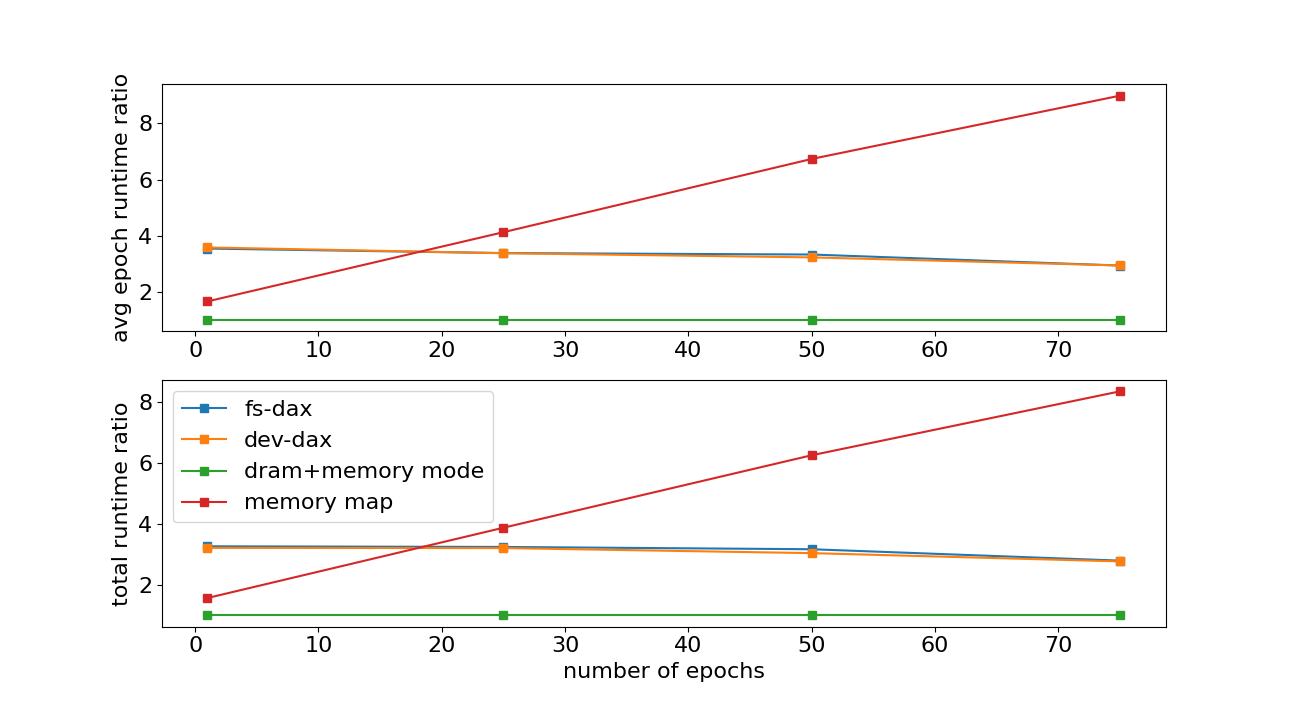}
\vspace{-1mm}
    \caption{\textbf{Ratios of runtimes (relative to DRAM)}. Note that \fsdax and \devdax are getting closer to DRAM speed as the amount of memeory increases.}
\vspace{-5mm}
    \label{fig:relative_ratios}
\end{figure}

The total runtime of our experiments, which can be seen in Figure~\ref{fig:total_runtimes}, shows that using \pmem is slower than DRAM by a factor of 1:3. While this is expected as the latency of \pmem to DRAM is also 1:3, we note that this gap is shrinking as the amount of memory used increases (as seen in Figure~\ref{fig:relative_ratios}). We also note that the ability to solely use DRAM is a rare case. In fact, in our simple experiments, we already ran into the case where the posterior was larger than DRAM capacity, and we were only able to produce results for the 470GB posterior by using \pmem (in Memory mode). In the case where the posterior cannot fit into DRAM, other \pmem configurations gain an advantage as \pmem in Memory mode is volatile, meaning all data must be serialized in order to persist. We do not report serialization costs in our experiments.

A large portion of time is used to allocate memory. Persistent \pmem configurations pay a steep penalty for crash-consistent heap allocation (orders of magnitude more time than on DRAM or memory mapping).  However, once the memory is allocated, using it obeys the 1:3 speed ratio as can be seen in Figure~\ref{fig:avg_epoch_runtimes}. This figure reports the average runtime of a single epoch of training. We note that while slower, persistent \pmem configurations combine computation and checkpointing into a single operation. After a successful write, the written data can be flushed from the caches and made persistent. If configured in Memory mode, a pessimistic user would have to serialize to disk after every computation in order to match the safety of using \fsdax or \devdax.

All storage methods were significantly faster than traditional memory mapping. We used the memory mapping functionality of NumPy~\cite{harris2020array} in our experiments, and we note that memory mapping does not scale as the memory size increases. To make matters worse, we did not explicitly flush the buffer after each write,. Therefore, the reported performance of our memory mapped experiments uses DRAM caching, and would be significantly worse if caching was disabled.

We did not notice a statistically significant different between \fsdax and \devdax performance. Even though \devdax was slightly faster, we suggest using \fsdax for greater system-wide flexibility via the mounted filesystem. Additionally, these modes also support a crash-consistency policy that use software undo-logging to protect against crashes or machine resets during writes. We did not enable this feature in our experiments in order to provide a fair comparison to existing storage methods (which do not have protected write operations).  Optimizing persistent memory transaction support is out of the scope of this paper.

One important behavior is the stability of \fsdax and \devdax. As the memory consumption increases, we observed large standard deviations in the runtime of memory mapping and Memory mode. This is a result of cache misses occurring in their implementation: NumPy will cache rows of the memory mapped file and evict data upon missing with a full cache. Likewise, Memory mode uses DRAM as a cache for the data, and when the memory size grows larger than DRAM capacity, the full DRAM cache starts evicting data on misses. Due to the nature of updating the posterior using random access, and our updates writing to columns of a matrix, each cache miss (when the cache is full) will induce a future cache miss upon the next write since rows are stored, by default, in row-major order in NumPy arrays.



\section{Future work}
In the future, we would like to further refine our posterior approximation. Currently our posterior is controlled via the number of samples to store, which is a hyper-parameter of the method. In general, using SGD iterates assumes that as the model trains, the iterates converge into high-probability areas of the posterior. This means that early iterates should be discarded while later iterates are useful. Where this boundary exists is unclear, and is something we wish to further explore.

Additionally, our experiments so far are using the MNIST dataset that can easily be stored in DRAM. Other, larger datasets are also good targets to store on \pmem to speed up data loading and preprocessing. However, if using Memory Mode, storing the data now competes with additional computation like the posterior for the DRAM cache. Further experimentation is needed to show the limits of \pmem Memory Mode in comparison to persistent configurations like \fsdax and \devdax.

Another avenue of future work is to explore more robust posterior approximations. While Maddox~\emph{at al}~\cite{maddox2019simple} provide a robust solution, their approach uses a single well-known distribution to approximate an arbitrary probability distribution. With \pmem accelerating the computation and providing large capacity memory, more complicated approximations are possible.

Finally, we wish to explore using \pmem in persistent mode with a small DRAM cache. Currently writes into all storage is random access and dominated by writing into columns of $\hat{D}$. In fact, \pmem and other storage methods have better sequential write performance. We could take advantage of this by storing a few local updates in DRAM and then writing the cache to storage once the cache is full.

\section{Conclusion}
In conclusion, \pmem is an incredibly usefull technology which will revolutionize modern computing. For data intensive applications, having access to memory which is fast, persistent, and high capacity is groundbreaking. In our paper we demonstrate that by using \pmem, we can accelerate important learning tasks such as posterior approximation without sacrificing runtime or precision.

\bibliographystyle{IEEEtran}
\bibliography{IEEEabrv, main}

\end{document}